\documentclass[runningheads]{llncs}
\usepackage{amssymb,mathtools}
\usepackage{graphicx}
\usepackage{color}
\def\Bh{\mbox{$\mathbf{h}$}}
\def\Be{\mbox{$\mathbf{e}$}}
\def\Bep{\mbox{$\mathbf{z}$}}
\def\Bo{\mbox{$\mathbf{o}$}}
\def\By{\mbox{$\mathbf{y}$}}
\begin{document}
\title{Neural Poetry: Learning to \\Generate Poems using Syllables\thanks{This is a post-peer-review, pre-copyedit version of an article published in LNCS, volume 11730. The final authenticated version is available online at: \texttt{https://doi.org/10.1007/978-3-030-30490-4\_26}}}
%
%
\author{Andrea Zugarini\inst{1,2},
Stefano Melacci\inst{2}, \and
Marco Maggini\inst{2}}
\authorrunning{A. Zugarini, S. Melacci, M. Maggini}
%

\institute{
	DINFO, University of Florence, Italy \\ 
	\and
	DIISM, University of Siena, Italy\\
	\email{andrea.zugarini@unifi.it}, \email{\{mela,maggini\}@diism.unisi.it}\\
}

\maketitle              
\begin{abstract}
Motivated by the recent progresses on machine learning-based models that learn artistic styles, in this paper we focus on the problem of poem generation. This is a challenging task in which the machine has to capture the linguistic features that strongly characterize a certain poet, as well as the semantics of the poet's production, that are influenced by his personal experiences and by his literary background.
Since poetry is constructed using syllables, that regulate the form and structure of poems, we propose a syllable-based neural language model, and we describe a poem generation mechanism that is designed around the poet style, automatically selecting the most representative generations. The poetic work of a target author is usually not enough to successfully train modern deep neural networks, so we propose a multi-stage procedure that exploits non-poetic works of the same author, and also other publicly available huge corpora to learn syntax and grammar of the target language.
We focus on the Italian poet Dante Alighieri, widely famous for his Divine Comedy. A quantitative and qualitative experimental analysis of the generated tercets is reported, where we included expert judges with strong background in humanistic studies.
The generated tercets are frequently considered to be real by a generic population of judges, with relative difference of 56.25\% with respect to the ones really authored by Dante, and expert judges perceived Dante's style and rhymes in the generated text.


\keywords{Poem Generation \and Transfer Learning \and Language Models \and Recurrent Neural Networks \and Natural Language Generation.}
\end{abstract}

\section{Introduction}
\label{sec:intro}
Natural Language Generation (NLG) is a challenging problem that has drawn a lot of attention in the Natural Language Processing (NLP) community \cite{reiter2000building,subramanian2017adversarial}. NLG is crucial for multiple NLP applications and problems, such as dialogue systems \cite{wen2015semantically}, text summarization \cite{chopra2016abstractive}, and text paraphrasing \cite{hasan2016neural}.
Poem generation is an instance of NLG that is particularly fascinating for its peculiar features. Verses have precise structures, rhyme and meter that convey an aesthetic and rhythmic sound to the poetry. This expressive art of language is ancient and spread across all cultures in the world. 

Automatically creating poems requires a strong attention to both the content and the form. The machine has to capture the linguistic features that strongly characterize a certain poet, as well as the semantics of the poet's production, that are influenced by their personal experiences and by their literary background.
In the last few years, the machine learning community focussed on the problem of poem generation, proposing approaches that generate either English quatrains \cite{colton2012full,hopkins2017automatically,lau2018deep} or Chinese verses \cite{zhang2014chinese,yu2017seqgan,yi2017generating,yi2018automatic}.
Most of them are based on neural architectures that combine several modules, post-processing the final results to generate well-formed verses (we postpone to Section~\ref{sec:related_work} an in-depth description of related work). In order to cope with the lack of large scale data, these works usually do not try to mimic the style of a target poet, and they frequently consider the poetic production of several authors. Moreover, despite Italian poetry is one of the most significant and well known poetries, to the best of our knowledge it has not been the subject of studies in which neural approaches have been evaluated. 

In this paper, we propose a simple and effective neural network-based model to generate verses. 
We focus on the Italian language and, in particular, on Dante Alighieri, the Italian poet that authored the Divine Comedy \cite{alighieri1998divine}, the most important poem of the Middle Ages, widely considered as the greatest literary work in the Italian literature. Our model learns to generate tercets with Dante Alighieri's style by ``reading'' the Divide Comedy. 
The learning problem is tacked following Language Modeling tasks but, differently from what is commonly done by several authors, we consider syllables as input tokens (instead of words, generic n-grams, or characters, for example). This choice is motivated by the fact that poetry is constructed using syllables that regulate the form and structure of poems. For example, syllables play a crucial role in the context of meter and rhyme. Moreover, the use of sub-word information is even more useful in Italian that is a language with a rich morphology. 
Our model consists of a Recurrent Neural Network (LSTM \cite{sundermeyer2012lstm}) that outputs one syllable at each time instant, conditioned to the previously generated text. The model is trained on Dante's tercets, that are composed of triples of hendecasyllables, with a precise structure of the rhymes. Due to its syllable-based nature, the proposed model can capture several properties of the input language, and it has a large flexibility in terms of what it can generate. The latter feature requires attention when using the language model to generate new text. We take into account the key properties of the Divine Comedy, automatically selecting the generations that are closer to Dante's style.

Training neural models on poems from a single target author can lead to low generalization quality, due to the small size of the training data.
Moreover, the language used by poets from the middle ages can be significantly different from modern language, such as in the case of the Italian used by Dante Alighieri and nowadays Italian.
We exploit the basic consideration that even if the form of some words have changed over time, there are a number of inherent regularities at the syllable level that have not changed that much. We propose to pre-train the system with a large modern Italian corpus (PAISA' \cite{lyding2014paisa}), and to perform transfer learning towards the poetry domain. The transfer of information is performed in multiple steps, exploiting Dante's prose and other Dante's poems (i.e. all the poet's production), and finally training the model with the Divine Comedy.

Our experimentation shows that exploiting Italian corpora and the poet's production improves the perplexity of the poetry-related language model, allowing the system to better capture the language and contents of the Divine Comedy. We performed a qualitative analysis of the generated tercets, based on human evaluation, where we also asked the collaboration of expert judges with strong background in humanistic studies. The generated tercets are frequently considered to be real by a generic population of judges, and expert judges perceived Dante's style and rhymes in the generated text. As expected, evaluators emphasized how the semantics behind the generated verses are sometimes hard to appreciate since they do not convey enough emotion, suggesting that more structured models that integrate additional information about the author could be an interesting topic for future work. 

The paper is organized as follows. In Section \ref{sec:related_work} we describe related state-of-the-art approaches. Then, we introduce the proposed model and the generation mechanism in Section \ref{sec:model}. Section \ref{sec:experiments} includes experiments and a discussion on the obtained results, while, finally, we draw our conclusions in Section \ref{sec:conclusions}. 

\section{Related Work}
\label{sec:related_work}
The scientific literature includes several works on machines that are either programmed to generate poems or that approach the problem of poem generation using machine learning algorithms. Early methods \cite{colton2012full} rely on rule-based solutions, while more recent approaches focus on learnable language models. Language Modeling is the problem of predicting which word comes next, given a sequence of previous words. In the last few years, neural language models are the dominant class of algorithms applied to NLG.  
While Language Modeling was successfully addressed using feed-forward neural networks on a fixed window of words \cite{bengio2003neural}, in \cite{mikolov2010recurrent} a recurrent neural network approach proved to be preferable. As a matter of fact, several nowadays NLG approaches are based on recurrent nets \cite{wen2015semantically,chopra2016abstractive,hasan2016neural}.

Word-based language models usually require large vocabularies to store all the (most frequent) words in huge textual corpora, and, of course, they cannot generalize to never-seen-before words. Some other approaches tried to overcome this issue, exploiting sub-word information.
A character-level solution was proposed in \cite{hierachicalcharlm}, while other authors \cite{miyamoto2016gated} combine word embeddings with character-level representations. It has been shown \cite{marra2018unsupervised,akbik2018contextual} that character-based models can be adapted to produce powerful word and even context representations that capture both morphology and semantics.
Sub-word information is very important in poetry, since it represents a crucial element to capture the ``form'' of a poem.

The first approach that proposes a deep learning-based solution to poem generation is described in \cite{zhang2014chinese}, where the authors combined convolutional and recurrent networks to generate Chinese quatrains. 
Then, a number of approaches focussing on Chinese poetry were proposed. In particular, a sequence-to-sequence model with attention mechanisms was proposed in \cite{yi2017generating} and \cite{wang2016chinese}.
In \cite{yu2017seqgan} the authors extend Generative Adversarial Networks (GANs) \cite{goodfellow2014generative} to the generation of sequences of symbols, exploiting Reinforcement Learning (RL). They consider the GAN discriminator to be the reward signal of a RL-based generator, and, among a variety of tasks, Chinese quatrains generation is also addressed.
Another RL-based approach is proposed in \cite{yi2018automatic}, where two networks learn simultaneously from each other with a mutual RL scheme, to improve the quality of the generated poems.

In the context of English poem generation, transducers were exploited to generate poetic text \cite{hopkins2017automatically}. Meter and rhyme are learned from characters by cascading a module that focusses on the content and a weighted state transducer that explicitly models the form of the generation.
Differently, the more recent Deep-speare \cite{lau2018deep} combines three neural modules, sharing the same character-based representation, to generate English quatrains. These models consist in a word-level language model fed with both word and character representations, a network that learns the meter, and another net that identifies rhyming pairs. At the end, generations are selected after a post-processing step that picks the best quatrains combining the output of the three modules.  
We notice that the authors of \cite{lau2018deep} exploited a collection of poems from several authors in order to train the model.

Following the intuition of working with syllables, our solution is simpler and in the case of Italian poetry, as we will show in Section~\ref{sec:experiments}, it generates tercets not only with the proper form, but also resembling the style of the selected target author. 

\section{Model}
\label{sec:model}
The main module of the proposed model consists of a syllable-based Language Model, also referred to as \textsc{sy-LM}, that processes a sequence of syllables. In order to handle the input data, we have to convert the available text into a sequence of syllables, i.e., we have to segment the input text into words, and then to split words into syllables. Since we focus on the Italian language, we implemented a module that follows the most common Italian hyphenation rules that, apart from rare exceptions, correctly divides words into syllables (the same procedure could have been followed for other languages, English included). 

We focus on data from Dante Alighieri's Divine Comedy, that is composed of a set of tercets (i.e., three verses). Each tercet is converted into a sequence of tokens (syllables) $x_{1}, \cdots, x_{T}$ belonging to the syllable dictionary $V_{s_y}$. We removed the punctuation and introduced some special tokens: word-separator $<$sep$>$, begin-of-tercet $<$go$>$, end-of-verse $<$eov$>$, end-of-tercet $<$eot$>$.

For each time instant $t$, \textsc{sy-LM} outputs the probability   
\begin{equation*}
	\hat{y}_{t} = p(x_t |  x_{1}, \cdots, x_{t-1} ) 
\end{equation*}
for all $x_t \in V_{s_y}$. If we indicate with $\hat{\By}_{t}$ the vector with the probabilities associated to all the vocabulary elements, the model yields the syllable associated with the highest probability.

We follow the classic setting of neural network-based language models: each element of the vocabulary is encoded into a 1-hot representation of size $|V_{s_y}|$, and the system associates it to a latent dense representation that is learned jointly with the other model parameters.
Such dense representations, also referred to as ``syllable embeddings'', are collected row-wise in matrix $W_{s_y} \in \mathbb{R}^{|V_{s_y}| \times d}$.
Each token $x_{t}$ of the input tercet is then mapped into its syllable embedding $\Be_{t} \in \mathbb{R}^d$, that is the row of $W_{s_y}$ associated to $x_{t}$. 
In detail, we have,
\begin{equation*}
	\Be_{t} =  W_{s_y} \cdot \mathbf{1}(x_{t}) \ ,
\end{equation*}
where $\mathbf{1}(\cdot)$ is a function returning the 1-hot column vector that has $1$ in the position associated to the vocabulary index of its argument.
It is important to notice that since $V_{s_y}$ is the set of all syllables (and a few special tokens), its cardinality is smaller than traditional word-based vocabularies, therefore the embedding matrix $W_{s_y}$ has a significantly smaller number of elements than what usually happens in the case of word-level representations. 

The sequence of syllable embeddings of the input tercet is provided as input to a recurrent neural network, one element at each time step. 
The internal state of the recurrent network at time $t$ is indicated with $\Bh_t$, and it is computed by updating the previous state using the current syllable embedding,
\begin{equation}
	\Bh_{t} = r(\Be_{t}, \Bh_{t-1}) \ ,
\end{equation}
where $r(\cdot,\cdot)$ is the state-update function of the network.
We selected LSTMs as recurrent neural model, due to their good results in language modeling \cite{sundermeyer2012lstm}.

A projection layer (weights $W$, bias $b$, activation $\sigma$ -- that we set to the hyperbolic tangent) transforms $\Bh_t$ into a $d$-sized vector $\Bep_t$, and a dense layer followed by the softmax activation function computes the probability distribution $\hat{\By}_{t}$,
\begin{eqnarray}
	\Bep_t = \sigma(W  \Bh_t + b)	 \label{eq:proj}\\ 
	\Bo_t = W_{s_y}'  \Bep_t \label{eq:output}\\ 
	\hat{\By}_t = \texttt{softmax}(\Bo_t) \label{eq:softmax} \ .
\end{eqnarray}
Notice that the dense layer of Eq. (\ref{eq:output}) shares its parameters with the syllable embedding matrix $W_{s_y}$ (being $'$ the transpose operator), ulteriorly reducing the number of learnable parameters of the model.

\begin{figure}[!h]
	\centerline{\includegraphics[scale=0.45]{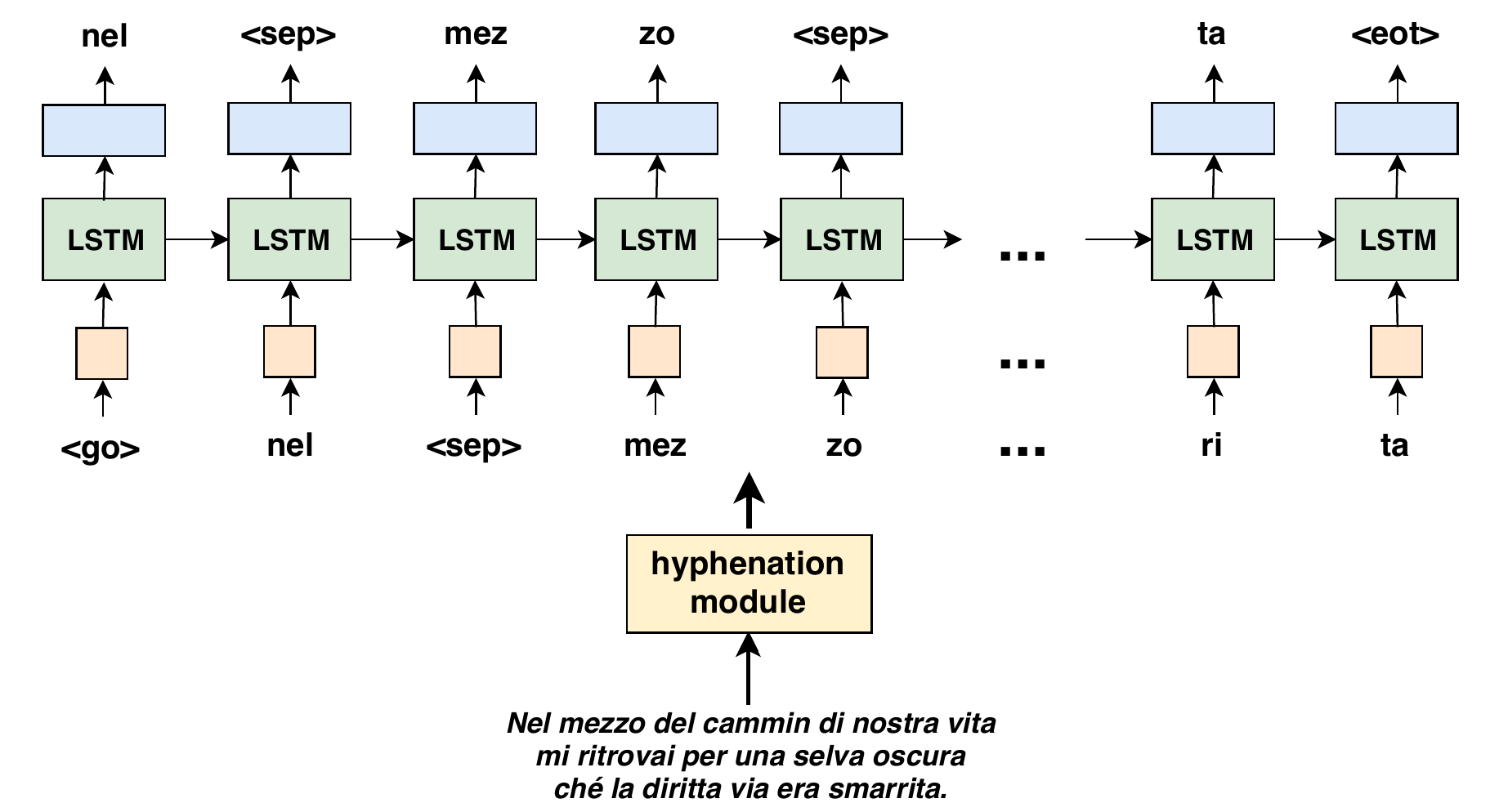}}
	\caption{Sketch of the Syllable LM. Input tercets are first pre-processed by a module that splits words into syllables and introduces some special tokens: word-separator $<$sep$>$, begin-of-tercet $<$go$>$, end-of-verse $<$eov$>$, end-of-tercet $<$eot$>$. Orange blocks are syllable embeddings, each blue block depicts the network of Eq. (\ref{eq:proj}-\ref{eq:softmax}).  The system predicts the next syllable, i.e. the one associated to the largest probability in $\hat{\By}_t$.}
	\label{fig:model}
\end{figure}

We train the \textsc{sy-LM} by minimizing the cross-entropy between each $\hat{\By}_t$ and the ground truth from the Divine Comedy, thus pushing toward $1$ the element of $\hat{\By}_t$ associated to the $t$-th syllable of the current tercet in the Divine Comedy. We measure the model performance in terms of perplexity (PPL), as commonly done in language modeling approaches \cite{mikolov2010recurrent}.
An illustration of the entire model is presented in Fig. \ref{fig:model}.

\subsection{Multi-Stage Transfer Learning}
\label{multi}
Learning from the Divine Comedy (or, more generally, from a single poem or from all the poems of a given target author) might not be enough to provide strong generalization skills to neural language models ($\approx 4,000$ tercets). For this reason, we follow a multi-stage training procedure that consists in sequentially pre-training our model with related data, before training it on the Divine Comedy. We want the model to deeply grasp most of the syntax and grammar of the Italian language, independently from the considered author, so that we pre-train the network using publicly available large Italian corpora (see Section \ref{sec:experiments}). Dante Alighieri lived in the middle ages, and he wrote the Divine Comedy in Tuscan/Florentine Italian dialect of that time, giving a strong contribute in creating the currently standard language for Italy. His language is slightly different from modern Italian, including some old-fashioned words and forms not used anymore. Word-level models are likely to fail due to the unavoidable changes in the vocabulary when moving from modern Italian to Dante's Italian. Differently, our syllable-based vocabulary is flexible enough to be transferred to related data. The transition toward the Divine Comedy can be made smoother by performing a further pre-training step using all Dante's production (poems and prose), thus allowing the network to get more information on the main linguistic features of the author. Finally, we train the model on the Divine Comedy.

\subsection{Poem Generation Procedure}
\label{gen}
Once \textsc{sy-LM} has been trained, it is directly exploited to generate new samples, i.e., new tercets.
We start with $\Bh_{0}$ set to zeros, and we feed the system with the $<$go$>$ input symbol, iteratively sampling the next token to generate. We follow a Monte Carlo sampling procedure as done in \cite{yu2017seqgan}. We keep sampling and generating tokens until the end-of-tercet symbol ($<$eot$>$) is generated or the number of syllables reaches a fixed maximum limit (75 in our experiments). Thanks to the randomness in the multinomial sampling, the system can generate multiple different sequences sampled from the distribution learned from the training data.

We generate a batch of tercets (2,000 in our experiments), and we assign a score $R(x) \in \mathbb{R}$ to each tercet $x$ of the batch. Those tercets with highest scores are selected among all the generated ones (only the top-scored generation, if the goal is to generate a single tercet).  
$R(x)$ is the average of $4$ different scores, $R_1(x), \ldots, R_4(x) \in \mathbb{R}$ that are based on known properties of the author of the Divine Comedy, in terms of form and language. In particular, tercets are composed of three hendecasyllables, with chained rhyming scheme (``ABA'' -- the first tercet is paired with the last one), and the words produced by the syllable-based generation must belong to the vocabulary used in the Divine Comedy.
The first score penalizes non-tercet-like generations,
\begin{equation}
	R_1(x) = -\texttt{abs}(|x| - 3) + 1 \ ,
\end{equation}
where $|x|$ indicates the number of verses in the tercets and \texttt{abs} is the absolute value function. 
Differently, $R_2(x)$ promotes sequences with verses in $x$ that follow an hendecasyllabic meter. Since our model is based on syllables, it is easy to count the number of syllables in a generated verse $v$, and we define $R_2$ as follows,
\begin{equation}
	R_2(x) = -\sum_{v \in x}(\texttt{abs}(|v| - 11)) + 1  \ .
\end{equation}
The chained rhyming scheme is measured by $R_3(x)$,
\begin{equation}
	R_3(x) =
	\begin{cases}
		1,& \text{if $(v_1, v_3)$, $v_1,v_3 \in x$ are in rhyme} \\
		-1,             & \text{otherwise}
	\end{cases} \ ,
\end{equation}
where a positive score is given when a tercet has first verse $v_1$ in rhyme with the third one $v_3$.
Since the generated $x$ is actually a sequence of syllables, words are identified by merging syllables until the word-separator token $<$sep$>$ is predicted. In order to avoid the generation of words that are far from the poet's style -- that is pretty unlikely in our experience --, we assign a small positive contribute $a$ to words in $x$ that belong to the vocabulary of the Divine Comedy. Formally,
\begin{eqnarray}
	R_4(x) = \sum_{w \in x}f_w(x) \label{eq:r_4},& &
	f_w(x_i) =
	\begin{cases}
		a,& \text{if $w \in V$} \label{eq:f_4}\\ 
		-b,             & \text{otherwise}
	\end{cases}
\end{eqnarray}
where $w$ indicates a word in tercet $x$. In the experiments $a$ was set to $0.05$ and $b$ to $1$ to strongly discourage not valid words.

\section{Experiments}
\label{sec:experiments}
We performed several experiments to assess the quality of the \textsc{sy-LM}, reporting both quantitative and qualitative results. We considered multiple data sources ($i.,\ ii.,\ iii.$ below), following the multi-stage learning procedure of Section \ref{multi}. The core of this work is the Divine Comedy, the most important Dante Alighieri's contribution. 

\noindent ($i.$) {The Divine Comedy} (\textbf{DC}). It is a poem composed of 100 ``cantos'' organized into three \emph{cantiche}. Each canto is a poem with a variable number of tercets  also known as ``Dante's tercet''.
\textsc{sy-LM} was trained on $3768$ tercets and evaluated on a test set of $472$. We also kept a validation set of $471$ to set the network hyper-parameters. Overall, there are about $180$k syllables in the Divine Comedy.

\noindent ($ii.$) {Modern Italian Dataset (\textbf{PAISA'}).}
We exploited {PAISA'},\footnote[1]{\url{http://www.corpusitaliano.it/en/contents/paisa.html}} a large corpus of Italian web texts. We considered a portion of $200$k documents, consisting of about $836$k sentences with more than $67$M syllables.

\noindent ($iii.$) {Dante's Production (\textbf{DP}).}
We collected most of Dante's known non-latin prose and poetry manuscripts. In particular, we gathered all the text from \emph{Convivio}, \emph{Le rime} and \emph{La vita nuova}, collecting overall $1752$ sentences ($\sim157$k syllables) for prose and  $2727$ verses ($\sim48$k syllables).

In order to select the hyper-parameters of the neural architecture we measured the perplexity (PPL) of several configurations on the validation set taken from the \textbf{DC} corpus.  
We found that the best performing size $d$ for the syllable embeddings was $300$, whereas the best size of the state of the LSTM was $1024$. State neurons were dropped out \cite{srivastava2014dropout} with probability $0.3$.
The size of $V_{s_y}$ was set to $1884$, including all the syllables in the Divine Comedy and the special tokens. When pre-training on \textbf{PAISA'} and then refining on \textbf{DP} (Section \ref{multi}), we kept a small validation set to decide when to early stop the learning procedure, and different batch sizes and learning rates have been validated. Best results occurred with batch size $32$ and learning rate of $0.001$. 

\subsection{Results}
We experimented the transfer learning procedure of Section \ref{multi}, evaluating the impact of the different data sources. In Table \ref{tab:ppl_res} we report our results (PPL) on both validation and test set data.
As expected, the model benefits from pre-training on additional data. In particular, the most significant improvement is given when pre-training on \textbf{PAISA'}, showing that there is a positive transfer of information from modern Italian to Dante's language. 
Moreover, despite the quantity of data in \textbf{DP} is still rather small, we can see further improvements when other Dante's productions are used to pre-train the model.

\begin{table}
	\caption{Perplexity on validation and test set data from the Divine Comedy, pre-training (or not) the model using multiple data. $A\rightarrow B $ means that we train on data $A$ first, and then we train on data $B$. }	\label{tab:ppl_res}
	\centering
	\begin{tabular}{c|cc}
		\hline
		\textbf{Datasets} & \textbf{Val PPL} &\textbf{Test PPL}\\ \hline
		\textbf{DC} &  12.45 &  12.39 \\
		\textbf{PAISA' $\rightarrow$ DC} & 10.83 & 10.82\\
		\textbf{DP $\rightarrow$ DC} & 11.95 & 11.74\\ 
		\textbf{PAISA'$\rightarrow$ DP $\rightarrow$ DC} & \textbf{10.63} & \textbf{10.55}\\ \hline
	\end{tabular}
\end{table}

The quality of the generated tercets has been assessed by human judges in two different evaluations.
In the first test, we involved $13$ graduate and not graduated students, mostly from humanistic degrees. We refer to them as ``non-expert'' judges, since they were not specialized in Dante's production, but very well aware of the author and of the Divine Comedy.
They were asked to judge if a given tercet was authored by Dante Alighieri or not (i.e., generated by \textsc{sy-LM}). Each judge evaluated $10$ tercets, $5$ of which were from Dante and $5$ generated by our model.
In Table \ref{tab:test_non_exp_0} we report the number of times (percentages) that tercets from a certain population were judged to be authored by Dante. It is clear that, given the humanistic background of the evaluators, judgements are rather thoughtful, however our generated tercets are considered as real almost half of the times of ones from Dante, with a relative difference of $56.25\%$.
\begin{table}[!ht]
	\caption{The number of times (percentages) that tercets from either \textsc{sy-LM} or Dante Alighieri (\textsc{Poet}) are judged to be authored by Dante (i.e., they were marked as ``real''). Our model is considered to be realistic almost half of the times of real Dante's production.}	\label{tab:test_non_exp_0}
	\centering
	\begin{tabular}{l|r}
		\hline
		\textbf{Generator} &    \textbf{Real-Mark} \\ \hline
		\textbf{sy-LM} &  28\% \\
		\textbf{Poet} &  64\% \\ \hline
	\end{tabular}
\end{table}

\begin{figure}[!ht]
	\centerline{\includegraphics[scale=0.35]{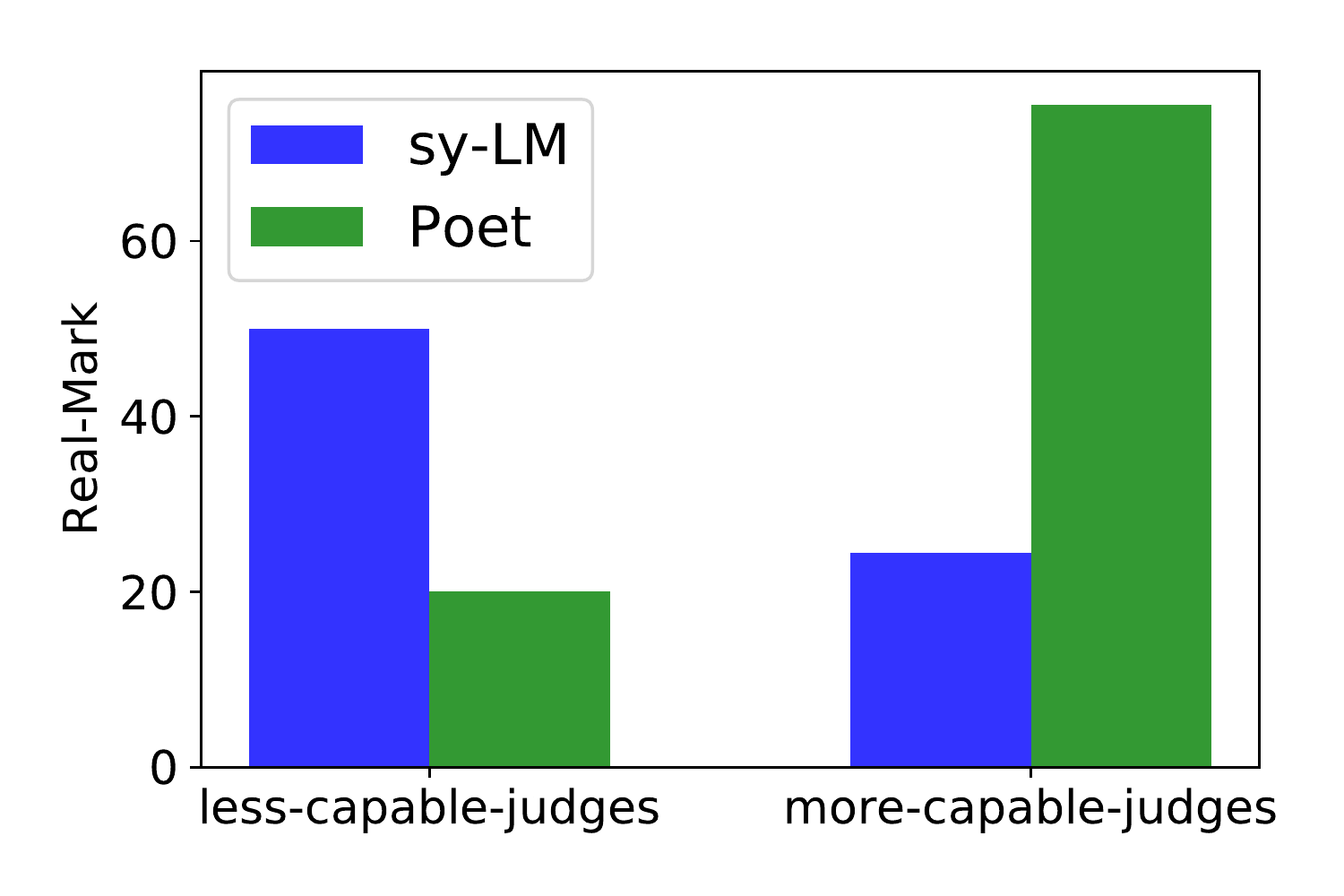}}
	\caption{Results of Table \ref{tab:test_non_exp_0} further divided into two groups: judges that are less capable of recognizing real Dante's tercets and the other ones.}	\label{fig:eval_groups}
\end{figure}

We can further analyze this result by distinguishing between those judges that were less capable of identifying real Dante's tercets (marking them to be real less than $50\%$ of the times) and the other ones. In Figure \ref{fig:eval_groups} we can observe that the ``less-capable judges'' were even more attracted by \textsc{sy-LM} than by real Dante's tercets. Since these judges better represent the average population of users, this result suggests that \textsc{sy-LM} is very positively perceived.
On the other hand, more capable evaluators are less frequently fooled by  \textsc{sy-LM} with a relative difference of $\approx 67\%$ from Dante. 

In another experiment, we involved $4$ expert judges with academic experiences on Dante Alighieri's production. 
Each expert evaluated $20$ tercets, scoring (from $0$ to $5$) different properties of each of them: \emph{emotion}, \emph{meter}, \emph{rhyme}, \emph{readability} and adherence to the \emph{author's style}. In particular, $10$ tercets were generated by \textsc{sy-LM} and $10$ were extracted from the Divine Comedy. Judges were not aware of how tercets were distributed.
We report the test results in Table \ref{tab:test_exp}. Dante's tercets are better scored, of course, however, we observe a good evaluation of the quality of the rhymes produced by  \textsc{sy-LM}. Considering that judges know very well Dante Alighieri, it is interesting to see that they are experiencing some of the author's style in the generated tercets. Evaluators emphasized how the semantics behind the generated verses are sometimes hard to appreciate since they do not convey enough emotion, that is the motivation behind the lower scores on the first two columns of Table \ref{tab:test_exp}. Finally, judges applied very strict criteria in evaluating the meter, giving low scores whenever a small incoherence with Dante's meter was apparently detected, even if they reported that it was not far from the ideal case.
\begin{table}[!ht]
	\caption{Experts evaluations restricted to tercets generated by \textsc{sy-LM}. Votes vary from $0$ to $5$. The average rate is also reported. For comparisons, in the last line we also report the average rate in the case of Dante's real tercets (\textsc{Poet}).}\label{tab:test_exp}
	\centering
	\begin{tabular}{l|ccccc}
		\hline
		\textbf{} &  \textbf{Readability} &   \textbf{Emotion} &     \textbf{Meter} &     \textbf{Rhyme} &     \textbf{Style} \\ \hline
		\textbf{Judge 1} & 1.57 & 1.21 & 1.57 & 3.36 & 2.29\\
		\textbf{Judge 2} &	1.64 & 1.45 & 1.73 & 3.00 & 2.27\\
		\textbf{Judge 3} & 2.83 &	2.33 &	2.00	&4.17&	2.92\\
		\textbf{Judge 4} & 2.17	& 2.00 &	2.33 &	2.92 &	2.50\\
		\hline
		\textbf{Average}           & 2.04 &	1.73 & 1.90 & 3.37 &	2.49 \\
		\hline
		\hline
		{Poet (Average)}         & 4.34 &	3.87 &	4.45 &	4.50 & 4.34 \\
		\hline
	\end{tabular}
\end{table}

Finally, we report some examples of generated tercets in Table \ref{tab:generations}. The first three tercets were well rated by non-expert and also expert judges, while the last one was badly scored.  
\begin{table}[!ht]
	\caption{Examples of generated tercets. The last one (bottom right) never fooled the judges, whereas the first three tercets were marked as real Dante's tercets by $88.00\%$, $55.56\%$ and $45.45\%$ of the evaluators, respectively.}\label{tab:generations}
	\centering
	\begin{tabular}{c}
		\hline
		\emph{e tenendo con li occhi e nel mondo} \\
		\emph{che sotto regal facevan mi novo} \\
		\emph{che 'l s'apparve un dell'altro fondo}\\
		\\
		\emph{in questo imaginar lo 'ntelletto}\\
		\emph{vive sotto 'l mondo che sia fatto moto}\\
		\emph{e per accorger palude è dritto stretto}\\ \hline
	\end{tabular}
	\begin{tabular}{ccc}
		& & \\
	\end{tabular}	
	\begin{tabular}{c}
		\hline
		\emph{per lo mondo che se ben mi trovi}\\
		\emph{con mia vista con acute parole}\\
		\emph{e s'altri dicer fori come novi}\\	
		\\
		\emph{non pur rimosso pome dal sospetto}\\
		\emph{che 'l litigamento mia come si lece}\\
		\emph{che per ammirazion di dio subietto}\\ \hline
	\end{tabular}
\end{table}

\section{Conclusions}
\label{sec:conclusions}
We presented a syllable-based language model for poem generation, that was applied to generate tercets. The proposed model is general, and we studied it in the context of Italian language and, in particular, in Dante Alighieri's Divine Comedy. 
Despite its simplicity and the lack of large-scale collections of data from the target author, our model produces tercets that are considered real by evaluators with humanistic background roughly half of the times of Dante's verses.
This is due to a scored generation mechanism that helps to keep Divine Comedy's meter and rhyme, and also due to a multi-stage training procedure that improves the quality of the content, exploiting all the poet's production and text in modern Italian. However, the outcome of the evaluation from expert judges clearly showed that, while the rhyme and style are positively captured by the model, the generations are still weak on meter and on conveying enough emotion.
In future work we plan to exploit the scoring criteria that we used in generating text to setup a Reinforcement Learning strategy. We are also interested in exploring more structured models that include additional information about the author, to improve the emotional quality of the generations.

\subsection*{Acknowledgments}
We thank Emmanuela Carb\'{e} and Elisabetta Bartoli for providing us Dante's data and for inviting several evaluators of our model. We would also like to thank Monica Marchi, Irene Tani, Maria Rita Traina and Simonetta Teucci, that helped us in evaluating the model.
This research was partially supported by QuestIT s.r.l. in the framework of the joint laboratory SAINLab.

\bibliographystyle{splncs04}
\bibliography{icann2019}

\begin{thebibliography}{10}
\providecommand{\url}[1]{\texttt{#1}}
\providecommand{\urlprefix}{URL }
\providecommand{\doi}[1]{https://doi.org/#1}

\bibitem{akbik2018contextual}
Akbik, A., Blythe, D., Vollgraf, R.: Contextual string embeddings for sequence
  labeling. In: Proceedings of the 27th International Conference on
  Computational Linguistics. pp. 1638--1649 (2018)

\bibitem{alighieri1998divine}
Alighieri, D., Sisson, C., Sisson, C., Higgins, D.: The Divine Comedy. Oxford
  University Press, Oxford University Press (1998)

\bibitem{bengio2003neural}
Bengio, Y., Ducharme, R., Vincent, P., Jauvin, C.: A neural probabilistic
  language model. Journal of machine learning research  \textbf{3}(Feb),
  1137--1155 (2003)

\bibitem{chopra2016abstractive}
Chopra, S., Auli, M., Rush, A.M.: Abstractive sentence summarization with
  attentive recurrent neural networks. In: Proceedings of the 2016 Conference
  of the NAACL: Human Language Technologies. pp. 93--98 (2016)

\bibitem{colton2012full}
Colton, S., Goodwin, J., Veale, T.: Full-face poetry generation. In: ICCC. pp.
  95--102 (2012)

\bibitem{goodfellow2014generative}
Goodfellow, I., Pouget-Abadie, J., Mirza, M., Xu, B., Warde-Farley, D., Ozair,
  S., Courville, A., Bengio, Y.: Generative adversarial nets. In: Advances in
  neural information processing systems. pp. 2672--2680 (2014)

\bibitem{hasan2016neural}
Hasan, S.A., Lee, K., Datla, V., Qadir, A., Liu, J., Farri, O., et~al.: Neural
  paraphrase generation with stacked residual lstm networks. In: International
  Conference on Computational Linguistics: Technical Papers. pp. 2923--2934
  (2016)

\bibitem{hopkins2017automatically}
Hopkins, J., Kiela, D.: Automatically generating rhythmic verse with neural
  networks. In: Proceedings of the 55th Annual Meeting of the Association for
  Computational Linguistics (Volume 1: Long Papers). vol.~1, pp. 168--178
  (2017)

\bibitem{hierachicalcharlm}
Hwang, K., Sung, W.: Character-level language modeling with hierarchical
  recurrent neural networks. In: Acoustics, Speech and Signal Processing
  (ICASSP), 2017 IEEE International Conference on. pp. 5720--5724. IEEE (2017)

\bibitem{lau2018deep}
Lau, J.H., Cohn, T., Baldwin, T., Brooke, J., Hammond, A.: Deep-speare: A joint
  neural model of poetic language, meter and rhyme  (2018)

\bibitem{lyding2014paisa}
Lyding, V., Stemle, E., Borghetti, C., Brunello, M., Castagnoli, S.,
  Dell'Orletta, F., Dittmann, H., Lenci, A., Pirrelli, V.: The paisa' corpus of
  italian web texts. In: 9th Web as Corpus Workshop (WaC-9)@ EACL 2014. pp.
  36--43. EACL (2014)

\bibitem{marra2018unsupervised}
Marra, G., Zugarini, A., Melacci, S., Maggini, M.: An unsupervised
  character-aware neural approach to word and context representation learning.
  In: International Conference on Artificial Neural Networks. pp. 126--136.
  Springer (2018)

\bibitem{mikolov2010recurrent}
Mikolov, T., Karafi{\'a}t, M., Burget, L., {\v{C}}ernock{\`y}, J., Khudanpur,
  S.: Recurrent neural network based language model. In: Eleventh annual
  conference of the international speech communication association (2010)

\bibitem{miyamoto2016gated}
Miyamoto, Y., Cho, K.: Gated word-character recurrent language model. In:
  Proceedings of the 2016 Conference on Empirical Methods in Natural Language
  Processing. pp. 1992--1997 (2016)

\bibitem{reiter2000building}
Reiter, E., Dale, R.: Building natural language generation systems. Cambridge
  university press (2000)

\bibitem{srivastava2014dropout}
Srivastava, N., Hinton, G., Krizhevsky, A., Sutskever, I., Salakhutdinov, R.:
  Dropout: a simple way to prevent neural networks from overfitting. The
  Journal of Machine Learning Research  \textbf{15}(1),  1929--1958 (2014)

\bibitem{subramanian2017adversarial}
Subramanian, S., Rajeswar, S., Dutil, F., Pal, C., Courville, A.: Adversarial
  generation of natural language. In: Proceedings of the 2nd Workshop on
  Representation Learning for NLP. pp. 241--251 (2017)

\bibitem{sundermeyer2012lstm}
Sundermeyer, M., Schl{\"u}ter, R., Ney, H.: Lstm neural networks for language
  modeling. In: Thirteenth annual conference of the international speech
  communication association (2012)

\bibitem{wang2016chinese}
Wang, Q., Luo, T., Wang, D., Xing, C.: Chinese song iambics generation with
  neural attention-based model. In: Proceedings of the Twenty-Fifth
  International Joint Conference on Artificial Intelligence. pp. 2943--2949.
  AAAI Press (2016)

\bibitem{wen2015semantically}
Wen, T.H., Gasic, M., Mrk{\v{s}}i{\'c}, N., Su, P.H., Vandyke, D., Young, S.:
  Semantically conditioned lstm-based natural language generation for spoken
  dialogue systems. In: Proceedings of the 2015 Conference on Empirical Methods
  in Natural Language Processing. pp. 1711--1721 (2015)

\bibitem{yi2017generating}
Yi, X., Li, R., Sun, M.: Generating chinese classical poems with rnn
  encoder-decoder. In: Chinese Computational Linguistics and Natural Language
  Processing Based on Naturally Annotated Big Data, pp. 211--223. Springer
  (2017)

\bibitem{yi2018automatic}
Yi, X., Sun, M., Li, R., Li, W.: Automatic poetry generation with mutual
  reinforcement learning. In: Proceedings of the 2018 Conference on Empirical
  Methods in Natural Language Processing. pp. 3143--3153 (2018)

\bibitem{yu2017seqgan}
Yu, L., Zhang, W., Wang, J., Yu, Y.: Seqgan: Sequence generative adversarial
  nets with policy gradient. In: Thirty-First AAAI Conference on Artificial
  Intelligence (2017)

\bibitem{zhang2014chinese}
Zhang, X., Lapata, M.: Chinese poetry generation with recurrent neural
  networks. In: Proceedings of the 2014 Conference on Empirical Methods in
  Natural Language Processing (EMNLP). pp. 670--680 (2014)

\end{thebibliography}
\end{document}